\documentclass{article}
\usepackage{spconf,amsmath,epsfig, hyperref,amsfonts}

\title{Efficient Subseasonal Weather Forecast using Teleconnection-informed Transformers}
\name{Shan Zhao, Zhitong Xiong, Xiao Xiang Zhu}
\address{Data Science in Earth Observation, Technical University of Munich (TUM), Ottobrunn, Germany}

\begin{document}
%
\maketitle
\begin{abstract}
Subseasonal forecasting, which is pivotal for agriculture, water resource management, and early warning of disasters, faces challenges due to the chaotic nature of the atmosphere. Recent advances in machine learning (ML) have revolutionized weather forecasting by achieving competitive predictive skills to numerical models. However, training such foundation models requires thousands of GPU days, which causes substantial carbon emissions and limits their broader applicability. Moreover, ML models tend to fool the pixel-wise error scores by producing smoothed results which lack physical consistency and meteorological meaning. To deal with the aforementioned problems, we propose a teleconnection-informed transformer. Our architecture leverages the pretrained Pangu model to achieve good initial weights and integrates a teleconnection-informed temporal module to improve predictability in an extended temporal range. Remarkably, by adjusting 1.1\% of the Pangu model's parameters, our method enhances predictability on four surface and five upper-level atmospheric variables at a two-week lead time. Furthermore, the teleconnection-filtered features improve the spatial granularity of outputs significantly, indicating their potential physical consistency. Our research underscores the importance of atmospheric and oceanic teleconnections in driving future weather conditions. Besides, it presents a resource-efficient pathway for researchers to leverage existing foundation models on versatile downstream tasks.
\end{abstract}
\section{Introduction}
\label{sec:intro}
The subseasonal forecast describes the trend of future atmospheric conditions in a few weeks ahead. Accurate weather forecasts play an important role in agriculture, aviation, disaster management, etc. The extended range forecast is challenged by the inherent chaotic nature of weather systems, which evolve through multi-scale processes and intricate nonlinear interactions. For example, $Ni\Tilde{n}o$ index, which originates from the central and eastern Pacific Ocean, could lead to heavy rains \cite{chand2017projected} and droughts \cite{allen201821st} in distant regions. Operational Numerical Weather Prediction (NWP) centers \cite{wedi2015modelling} provide global seasonal forecasts. Their advances benefit from sophisticated data assimilation algorithms, computing development, and improved understanding of physical processes. However, there remain challenges in physical process parameterization, uncertainty formulation, and the provision of physically consistent initial conditions for forecasts using observations \cite{bauer2015quiet}. In recent years, rapid progress has been made in the quality of deep learning (DL)-based weather forecasts. WeatherBench \cite{rasp2020weatherbench, rasp2023weatherbench} provides a benchmark dataset for data-driven medium-range weather forecasting. FourCastNet \cite{kurth2023fourcastnet} uses adaptive Fourier Neural Operators for accurate short to medium-range global predictions, and it achieves comparable predictability to the Integrated Forecasting System (IFS). Pangu-Weather \cite{bi2023accurate} takes a SWIN transformer \cite{liu2021swin} with Earth-specific positional bias to improve the 3D forecast and it achieves better forecast skills on targeting variables compared to IFS. However, these ML models fail to predict beyond 10 days lead. Besides, they cost considerable training efforts and their lack of physical modeling hurdles the trust of researchers in model results. 
\par
In this paper, we propose to consider teleconnections to encourage physical process modeling. Specifically, we select 16 climate indices whose temporal features are processed by inception modules with various kernel sizes. We leverage the pre-trained Pangu model with a lead time of 24 hours. After adapting 1.1\% of the model parameters, we achieve better spatial saliency and reduced error scores with the extended lead time of 336h. Our study presents a feasible strategy for researchers from diverse backgrounds to effectively utilize foundation models in their investigations.
\begin{figure*}[htb]
    \centering
    \includegraphics[width=1.8\columnwidth]{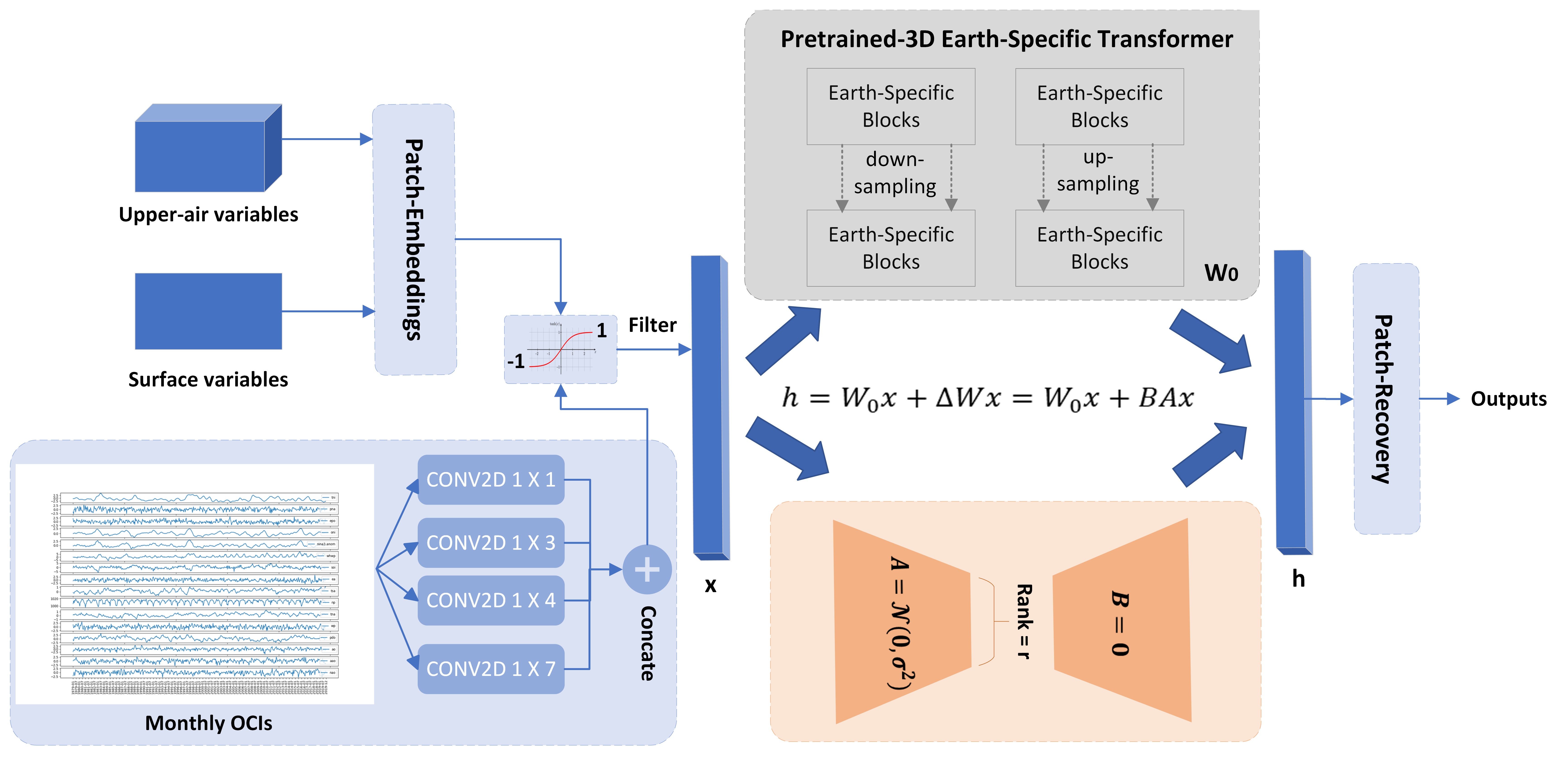}
    \caption{\small Workflow of teleconnection-informed transformers. The input signals are upper-air variables, surface variables, and OCIs. The OCIs are processed by the temporal module with various kernel sizes to filter the features of climate variables. The 3D Earth-Specific Transformers (Pangu) take the pre-trained weights at 24h lead time and use the LoRA strategy to achieve predictability over longer horizons.}
    \label{fig:workflow}
\end{figure*}
\section{Methods}
\label{sec:methods}
We formulate the subseasonal forecasts as a multi-modality video prediction task. The inputs are climate variables $\mathbf{X}_{\text{surface}} \in \mathbb{R}^{C_1\times H\times W}$, $\mathbf{X}_{\text{upper}} \in \mathbb{R}^{C_2\times H \times W\times Z}$, and Ocean Climate Indices (OCIs) $\mathbf{X}_{\text{oci}} \in \mathbb{R}^{C_3 \times L}$. In our global forecast task, $H$ is the longitudinal grids, $W$ is the latitudinal grids, $Z$ is the number of pressure levels, $L$ is the lags to the current time step $t$, and $C$ is the number of variables in each input. The target is to predict the future weather status $\mathbf{Y}_{\text{surface}}$ and $\mathbf{Y}_{\text{upper}}$ at time step $t+h$. $h$ is the horizon, i.e., how far ahead the model predicts the future. We use a temporal module to process OCIs, whose activation is between $+/-$1 and is regarded as driving signals of future weather patterns. The transformer-based model learns the attention across the different patches of the global measurements and updates its weights using a low-rank strategy \cite{hu2021lora}. The workflow is briefed in Figure \ref{fig:workflow}.
\subsection{Teleconnection-informed temporal filter}
Limited by computational time and memory, most DL models take a single frame as the input and regard the prediction as an image-image translation task. These models overlook the temporal information which is instrumental for capturing longer-term dynamics in weather patterns. Moreover, their input contains only the target variables, in which the system's complexity and various driving factors are ignored. Prapas et al. \cite{prapas2023televit} validated the effectiveness of using OCIs to improve forecast performance in complex Earth system. Inspired by this, we propose incorporating the OCIs as an additional model input. The OCIs temporally or spatially aggregated from measurements to a condensed 1D space are simple yet informative to indicate future global climate patterns \cite{hurrell1996influence, he2013multiresolution, lim2013impact}. 
\par
We use a multi-level feature extractor to capture information at various scales and complexities in OCIs. Specifically, to capture the daily, seasonal, semi-annual, and annual changes of Earth-related phenomena, we use inception modules \cite{szegedy2015going} with kernel sizes of (1,1), (1,3), (1,4), and (1,7). For example, the annual period is covered by a 3-layer Neural Network of combined kernel sizes of (1,7), (1,3), and (1,4). The output of each inception module is concatenated together to cover  movements of various types. The output of the inception model is fed to a hyperbolic tangent activation, functioning as a feature selector of the input atmospheric variables in the embedded feature space.

\subsection{Parameter efficient adaptation}
Earth observation data typically experience significant shifts due to sensory change, geographical difference, and temporal displacement \cite{zhao2022graph}. To adapt the pre-trained model to different tasks or domains, executing pre-training on benchmark datasets before fine-tuning deeply learned models for downstream becomes a feasible solution \cite{wang2022self}. Among the transfer learning methods, linear probing and full fine-tuning are the most general procedures. However, these methods require significant computational resources and can lead to overfitting and catastrophic forgetting \cite{kirkpatrick2017overcoming} issues. 
\par
To extend the prediction horizon of the model effectively, we use the parameters of the pre-trained Earth-specific 3D Transformers at the lead time of 24 hours as an initialization. Then, we implement a low-rank adaptation (LoRA) of foundation models. The core idea is that the learned over-parametrized models in fact reside on a low intrinsic dimension. The updated model weights are composed of a frozen pre-trained part, and the trainable part, which uses rank decomposition matrices. The two components are balanced by a hyperparameter $\alpha$. The forward path to extract the feature of a LoRA layer is 
\begin{equation} \small
    h=W_0x+\Delta Wx = W_0x +BAx,
\end{equation}
where $h$ is the output of the layer, $x$ is the input of the layer, $W_0$ is the frozen weight heritaged from the pre-trained model, $B$ and $A$ are trainable parameters, where $B \in \mathbb{R}^{d\times r}$ and $A \in \mathbb{R}^{r\times k}$. The rank $r << \min (d,k)$. We take random Gaussian initialization for $A$ and zeros for $B$. The task-specific parameters are updated during the backpropagation.
\begin{equation} \small
    \mathop{\max{_\theta}} \sum_{(x,y)\in \mathcal{Z}} \sum_{t=1}^{|y|} \log(p_{\phi_0+\Delta\phi(\theta)}(y_t|x,y_{<t})).
\end{equation}
\section{Experimental validation}
\label{sectionExperiments}
\subsection{Data}
We use two modalities of Earth observation data, including grided climate variables and time-series climate indices. We downloaded five years of the 5th generation of ECMWF reanalysis (ERA5) data. The selection of climate variables in our study follows the setting in \cite{bi2023accurate}. Specifically, five variables (geopotential, specific humidity, temperature, u-component and v-component of wind speed), each with 13 pressure levels (50hPa, 100hPa, 150hPa, 200hPa, 250hPa, 300hPa, 400hPa, 500hPa, 600hPa, 700hPa, 850hPa, 925hPa, and 1000hPa), and four variables (2m temperature, u-component and v-component of 10m wind speed, and mean sea-level pressure) at the surface level sampling at 00:00 UTC and 12:00 UTC are selected. The climate variables are normalized by the past 30-year atmospheric status. The teleconnection indices selected are Antarctic Oscillation Index, East Atlantic, North Atlantic Oscillation, Tropical Northern/South Atlantic, North Pacific, Pacific Decadal Oscillation, Eastern Pacific Oscillation, Pacific-North American, West Pacific, Southern Oscillation Index, Trans-Niño Index, Western Hemisphere Warm Pool, Arctic Oscillation, Niño 3, and Oceanic Niño Index, which are available at \cite{indices}. We incorporate OCIs from the 22 months preceding the current input time step. For any missing values within the time series, we employ a zero-filling strategy. Subsequently, the OCIs are normalized using their mean and standard deviation. We use 2015-2017 data for training, 2019 data for validation, and 2018 data for testing.

\subsection{Evaluation}
To evaluate the model performance, we compute the latitudinal-weighted Root Mean Square Error (RMSE) and the latitudinal-weighted Anomaly Correlation Coefficient (ACC).
The equation to compute RMSE of the variable $v$ at prediction time $t$ is
\begin{equation} \small
    \text{RMSE}(v,t) = \sqrt{\frac{\sum^{W}_{i=1}\sum^{H}_{j=1}L(i)(\hat{\mathbf{Y}}^v_{i,j,t}-\mathbf{Y}^v_{i,j,t})^2}{W\times H}},
\end{equation}
\begin{equation} \small
    \text{ACC}(v,t) = \frac{\sum_{i,j}L(i)\hat{\mathbf{Y}}^{'v}_{i,j,t}-\mathbf{Y}^{'v}_{i,j,t}}{\sqrt{\sum_{i,j}L(i)(\hat{\mathbf{Y}}^{'v}_{i,j,t})^2 \times \sum_{i,j}L(i)(\mathbf{Y}^{'v}_{i,j,t})^2}},
\end{equation}
where $L(i)$ is the weight at latitude $\phi_i$. $\mathbf{Y}'$ denotes the difference between $\mathbf{Y}$ and the climatology. However, the quality of weather forecasts is more than scores, we also qualitatively compared the spatial granularity of the results. We choose both DL-based models and numerical modeling-based models for comparison.
\begin{itemize}
    \item AR-Pangu \cite{bi2023accurate}. We use the released Pangu model with a lead time of 24 hours, the longest prediction horizon a single model can achieve, in an auto-regressive (AR) manner to forecast weather conditions 14 days ahead.
    \item Full finetune. We reproduce the Pangu model based on the pseudocode available at \cite{PanguWeathercode} and load 24-hour model weights, then we finetune Pangu to predict the 14-day ahead weather directly. 64 million parameters are updated.
    \item IFS \cite{wedi2015modelling}. IFS provides operational forecasts produced with the ECMWF. The IFS scores at 1.40625$^\circ$ reported at \cite{nguyen2023climax} are used for comparison.
\end{itemize}

\subsection{Results}
\begin{figure*}[htb]
    \centering
    \includegraphics[width=1.8\columnwidth]{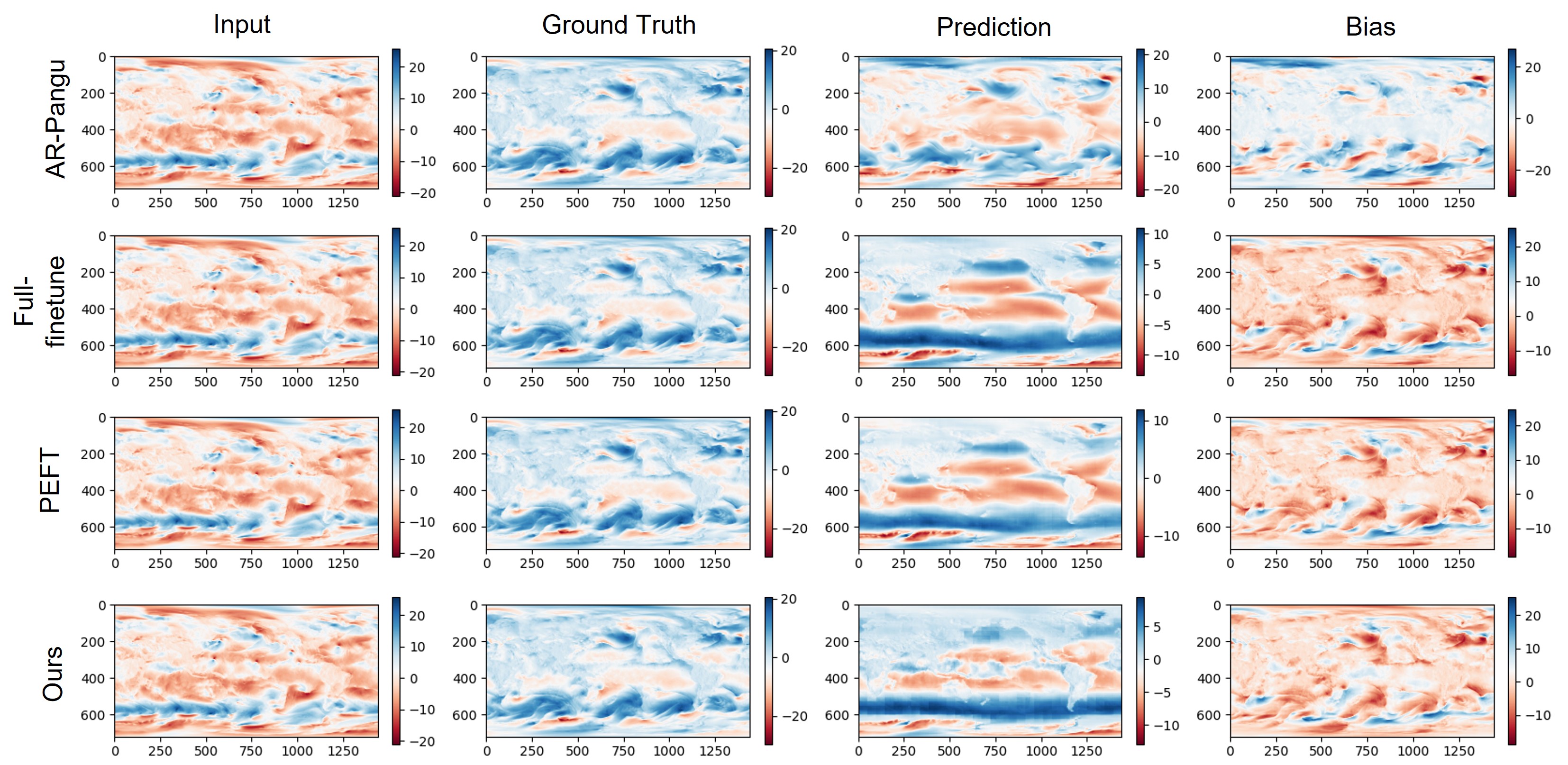}
    \caption{\small Example visualizations of subseasonal forecast of u10m by proposed method and baselines. From left to right are the input frame at 2018-10-01 00:00 UTC, target frame at 2018-10-15 00:00 UTC, prediction, and bias between the prediction and the target frame.}
    \label{fig:u10m}
\end{figure*}
The model is trained using the Adam optimizer \cite{kingma2014adam} at an initial learning rate of 2e-5 and a weight decay of 3e-6. We implement a multi-step scheduler, applying a decay factor of 0.5 every 15 epochs throughout a training period of 30 epochs. The temporal module is initialized by a truncated normal distribution with a mean of zeros and a standard deviation of 0.02. The input and output embedding layers are initialized by pretrained weights and later finetuned. For the linear layers, we adapt using LoRA where $r$ is 32, $\alpha$ is 16, and LoRA dropout rate of 0.1 to accommodate the high complexity inherent in EO data. Overall, this approach updates approximately 700k parameters.
\begin{table}[htb]
\small
    \centering
    \caption{\small The Latitudinal-weighted RMSE of proposed model and baselines on prediction tasks at 14 days ahead. The lower RMSE is favored for the better predictability of the model. The highest score is highlighted with bold text, and the second-best score is emphasized with underlining.\\}
    \begin{tabular}{c c c c}
    \hline
        Methods & T2M & Z500 & T850 \\ \hline
        Randomness & 3.75 & 1129.34 & 4.86 \\
        AR-Pangu (0.25$^{\circ}$) & 3.32 & 996.24  & 4.37 \\
        Full finetune (0.25$^{\circ}$) & \textbf{2.71} &  \textbf{832.33} &  \textbf{3.53}\\
        IFS (1.40625$^{\circ}$) & 3.30 & 1011.56 & 4.43\\
        Ours (0.25$^{\circ}$) &  \underline{3.17} & \underline{871.74} & \underline{3.91}\\ \hline
    \end{tabular}
    \label{tab:rmse}
\end{table}
\begin{table}[htb]
\small
    \centering
        \caption{\small The Latitudinal-weighted ACC of proposed model and baselines on prediction tasks at 14 days ahead. A larger ACC indicates a greater similarity between the predicted image and the ground truth image.\\}
    \begin{tabular}{c c c c}
    \hline
        Methods & T2M & Z500 & T850 \\ \hline
        Randomness & 0.98 & 0.93 & 0.94 \\
        AR-Pangu (0.25$^{\circ}$) & 0.98 & 0.95 & 0.95  \\
        Full finetune (0.25$^{\circ}$) & \textbf{0.99} & \textbf{0.96} & \textbf{0.97}\\
        IFS (1.40625$^{\circ}$) & 0.85 & 0.55 & 0.69 \\
        Ours (0.25$^{\circ}$) & \underline{0.98} & \textbf{0.96} & \underline{0.96} \\ \hline
    \end{tabular}
    \label{tab:acc}
\end{table}
\par
Table \ref{tab:rmse} and \ref{tab:acc} are the RMSE and ACC of variable surface temperature (T2M), geopotential at 500 hPa (Z500), and temperature at 850 hPa (T850). Additionally, we calculate these scores under a unique scenario where the output is simply replicated from the input, denoted as "Randomness" in the tables. This scenario serves as an upper bound of the error that a model could potentially reach. IFS demonstrates performance close to "Randomness". This result suggests that pixel-wise scoring is not an optimal metric for evaluating subseasonal weather forecast performance, considering IFS is widely regarded as the state-of-the-art NWP model. Although the fully fine-tuned model attains the best error scores, a closer examination of the outputs, such as in Figure \ref{fig:u10m}, reveals a noticeable smoothness. This smoothing effect leads to the elimination of certain important weather patterns, which indicates the lack of meteorological meaning and physical consistency. AR-Pangu, on the other hand, preserves spatial granularity but achieves poor forecast scores. Distinguishing from the above baselines, our proposed model shows a good compromise between the computation cost, error scores, and spatial saliency. 
\par
To assess the impact of the teleconnection-informed temporal module, we experiment with the LoRA strategy without OCIs as an input (denoted by PEFT). The impact of OCI in shaping future weather patterns is illustrated through the visualization of sample frames of the u-component of wind speed (u10m) on 2018-10-15 at 00:00 UTC, as depicted in Figure \ref{fig:u10m}. Incorporating the teleconnection index as a filtering mechanism enhances the spatial granularity of the model's output. This improvement suggests that our module is adept at capturing and preserving essential weather patterns, underscoring its potential for meteorological uses.
\section{Conclusion}
\label{sectionConclusion}
The advent of DL models has marked a significant shift in weather forecasting, enhancing predictive accuracy as evidenced by reduced error scores. However, skeptics are concerned about the absence of explainability and physical knowledge in the data-driven models. Additionally, despite their rapid inference, these models demand considerable computational resources for training and often exhibit limited generalizability across downstream tasks.
\par
Our proposed workflow introduces a temporal module that leverages various teleconnection indices as predictors of future weather conditions. By tailoring kernel sizes to align with atmospheric motion cycles, we observe improvements in spatial granularity, indicating the model's ability to retain crucial weather patterns. Furthermore, we employ a low-rank adaptation to reduce the number of parameters updated, thereby enhancing the possibility of using the pretrained model for various downstream tasks. Our method achieved reduced RMSE on T2M, Z500, and T850 for 0.15, 124.5, and 0.44, respectively, after adaptation, coupled with an increased ACC. In the future, we aim to extend to more varied tasks, such as predicting new variables, spatial downscaling, and improving performance on extreme events. Our goal is to not only improve the predictive accuracy but also to produce meteorologically and physically consistent predictions.

\small
\bibliographystyle{IEEEbib}
\bibliography{refs}

\end{document}